\documentclass[sn-mathphys-ay]{sn-jnl}

\usepackage{graphicx}%
\usepackage{multirow}%
\usepackage{amsmath,amssymb,amsfonts}%
\usepackage{amsthm}%
\usepackage{mathrsfs}%
\usepackage[title]{appendix}%
\usepackage{xcolor}%
\usepackage{textcomp}%
\usepackage{manyfoot}%
\usepackage{booktabs}%
\usepackage{algorithm}%
\usepackage{algorithmicx}%
\usepackage{algpseudocode}%
\usepackage{listings}%

\theoremstyle{thmstyleone}%
\theoremstyle{thmstyletwo}%

\theoremstyle{thmstylethree}%

\begin{document}

\title[Article Title]{Learning Global Object-Centric Representations via Disentangled Slot Attention}


\author[1]{\fnm{Tonglin} \sur{Chen}}\email{tlchen18@fudan.edu.cn}
\author[1]{\fnm{Yinxuan} \sur{Huang}}\email{yxhuang22@m.fudan.edu.cn}
\author[1]{\fnm{Zhimeng} \sur{Shen}}\email{zmshen22@m.fudan.edu.cn}
\author[1]{\fnm{Jinghao} \sur{Huang}}\email{jhhuang22@mfudan.edu.cn}
\author*[1]{\fnm{Bin} \sur{Li}}\email{libin@fudan.edu.cn}
\author[1]{\fnm{Xiangyang} \sur{Xue}}\email{xyxue@fudan.edu.cn}

\affil[1]{\orgdiv{School of Computer Science}, \orgname{Fudan University}, \orgaddress{\street{220 Handan Rd}, \city{Shanghai}, \postcode{200433}, \country{China}}}


\abstract{Humans can discern scene-independent features of objects across various environments, allowing them to swiftly identify objects amidst changing factors such as lighting, perspective, size, and position and imagine the complete images of the same object in diverse settings. Existing object-centric learning methods only extract scene-dependent object-centric representations, lacking the ability to identify the same object across scenes as humans. Moreover, some existing methods discard the individual object generation capabilities to handle complex scenes. This paper introduces a novel object-centric learning method to empower AI systems with human-like capabilities to identify objects across scenes and generate diverse scenes containing specific objects by learning a set of global object-centric representations. To learn the global object-centric representations that encapsulate globally invariant attributes of objects (i.e., the complete appearance and shape), this paper designs a \textit{Disentangled Slot Attention} module to convert the scene features into scene-dependent attributes (such as scale, position and orientation) and scene-independent representations (i.e., appearance and shape). Experimental results substantiate the efficacy of the proposed method, demonstrating remarkable proficiency in global object-centric representation learning, object identification, scene generation with specific objects and scene decomposition.}

\keywords{Global Object-Centric Representations, Object Identification,  Unsupervised Learning, Disentangled Learning }



\maketitle

\section{Introduction}\label{sec1}
Humans perceive the world in a compositional manner \citep{Lake_Ullman_Tenenbaum_Gershman_2017}, identifying visual concepts as intermediate representations to support scene understanding and high-level reasoning. Inspired by the human perception system, \textbf{O}bject-\textbf{C}entric \textbf{L}earning (OCL) aims to decompose a visual scene into visual concepts to uncover its structural information without any supervision \citep{10.1109/TPAMI.2023.3286184}.

Due to variations in object position, scale, pose, and illumination, individual objects can generate an infinite array of distinct images \citep{dicarlo2007untangling}. However, existing models have not paid much attention to the impact of these factors on visual concepts. While some OCL methods, such as AIR~\citep{eslami2016attend}, SPACE~\citep{lin2019space}, and GMIOO~\citep{yuan2019generative}, effectively disentangle object appearance, shape, size, and position, the resulting representations tend to be scene-specific and may not generalize to other contexts. Consequently, the learned object-centric representations display considerable variability across different scenes \citep{chen2024compositional}, greatly limiting their usability and robustness. In contrast, humans can discern scene-independent features of objects across various environments~\citep{karimi2017invariant}, allowing them to rapidly and accurately identify objects amidst these variable factors and imagine the complete images of the same object in diverse settings. To empower AI systems with human-like capabilities, we propose a disentangled slot attention algorithm designed to learn global, scene-independent, object-centric representations rather than local, scene-dependent ones.

Despite significant progress in unsupervised object segmentation of complex natural scenes, the generative capability of some existing OCL methods has been compromised. SLATE~\citep{singh2022illiterate} is the first to employ an auto-regressive transformer decoder to enhance object segmentation. OSRT~\citep{sajjadi2022object} introduces a highly efficient decoder called Slot Mixer to speed up the compositional rendering of NeRF-based methods. LSD~\citep{jiang2023object} utilizes a diffusion model to improve slot-to-image decoding quality. However, these methods combine the slots in the decoder for image reconstruction instead of decoding each slot independently, thus limiting their ability to generate images containing only a single object. In addition, DINOSAUR~\citep{Seitzer2022BridgingTG} adopts the self-supervised pre-trained ViT model DINO~\citep{caron2021emerging} to extract image features to improve the ability to segment complex scenes significantly. It only reconstructs image features rather than entire images for training, resulting in a loss of image generation capability. Furthermore, none of these methods can identify the object across scenes and generate scenes containing specific object identities. Therefore, the goal of this work is to equip object-centric models with the ability to identify objects and enhance their generative capacity while preserving their segmentation performance.

Similar to the work in this paper, GOCL \citep{chen2024compositional} can identify the same objects in different scenes through learnable global object-centric representations. GOCL focuses on calculating the similarity between the global object-centric representations (indicating the global invariant appearance and shape of the complete object) and the intrinsic attribute (i.e., appearance and shape) representation of the object (probably occluded) in the current scene through a patch-matching strategy, thereby predict the identity of each object (probably occluded) in the current scene. Therefore, it is only suitable for simple 2D scenes and cannot handle complex 3D scenes. It limits the potential of GOCL to generalize to real-world 3D scenes.

In this paper, we propose a novel object-centric learning method, called  \textbf{G}lobal \textbf{O}bject-centric \textbf{L}earning via \textbf{D}isentangled slot attention (GOLD). This method aims to identify objects across various scenes and generate diverse scenes containing specific objects by learning a set of global object-centric representations. GOLD comprises two key components: an \textit{Image Encoder-Decoder} module and a \textit{Global Object-Centric Learning} module. The \textit{Image Encoder-Decoder} module extracts the image features through ViT encoder DINO~\citep{caron2021emerging} and decodes the reconstruction of image features to the entire image through a VQ-VAE~\citep{van2017neural} decoder. On the other hand, the \textit{Global Object-Centric Learning} module focuses on learning global object-centric representations that encapsulate globally invariant attributes of objects (i.e., the complete appearance and shape). Specifically, it employs distinct encoding and decoding processes for distinguishing background from objects. Furthermore, it introduces a \textit{Disentangled Slot Attention} module, a disentangled variant of slot attention, which separates an object representation into extrinsic attribute representation and object identity representation. The former encompasses scene-dependent attributes (such as scale, position, and orientation), while the latter is used to select from a learnable set of global object-centric representations to compute intrinsic attribute representation, which represents scene-independent attributes (i.e., appearance and shape). Finally, the intrinsic representation and extrinsic representation are concatenated and input into a mixture-based decoder to reconstruct image features extracted by DINO.

We conduct experiments on four datasets. We show both qualitatively and quantitatively that in contrast to the prior art, the proposed model demonstrates remarkable proficiency in global object-centric representation learning, object identification, and scene generation with specific objects.

In summary, our main contributions are the following:
\begin{itemize}
    \item [1.] We propose GOLD, an object-centric learning method possessing the ability to identify objects through learning global object-centric representations.
    \item [2.] We introduce a \textit{Disentangled Slot Attention} module to convert the scene features into scene-dependent and scene-independent attribute representations for each object in the current scene;
    \item [3.] GOLD is the first object-centric learning method that learns global object-centric representation and can identify objects in complex 3D scenes;
    \item [4.] GOLD has outstanding scene-understanding capabilities on four datasets.
\end{itemize}

\section{Related Works}
\label{sec:related}
Object-centric representation learning methods are attracting increasing attention. Existing methods can be roughly categorized into two types based on the way object representations are extracted: non-Slot Attention and Slot Attention-based.

\textbf{Non-Slot Attention:} AIR~\citep{eslami2016attend} is an earlier representative method. It extracts representations of objects in sequence based on the attention mechanism. The representation of each object can be divided into three parts: presence, position, and appearance. GMIOO~\citep{yuan2019generative} improves AIR by handling occlusions between objects and models objects and the background separately. It infers the representations of objects by amortized variational inference and generates the whole scenes iteratively through an LSTM module. IODINE~\citep{greff2019multi} utilizes iterative variational inference to learn the representation of objects in the scene. SPACE~\citep{lin2019space} models the background with spatial mixture models and extracts the representations of foreground objects with parallel spatial attention. ROOTS~\citep{chen2021roots} divides a 3-dimensional scene into grid cells extending SPACE and then estimates the size and position of each object in 3-dimensional space. The representation of each object is obtained by encoding the features of objects over multiple viewpoints. GENESIS~\citep{engelcke2019genesis} models the interrelationship between objects in an autoregressive manner. GENESIS-V2~\citep{engelcke2021genesis} improves the performance of GENESIS by predicting mask attention of objects with the Instance Colouring Stick-Breaking Process. SQAIR~\citep{kosiorek2018sequential} extends the discovery and propagation modules in AIR to discover and track objects throughout the sequence of frames. In addition, SQAIR can generate future frames conditioning on the motion of objects in the current frame. These methods, similar to our work, can reconstruct images of individual objects and entire scenes. However, their effectiveness is limited to simple synthetic datasets, and they lack the capability for object recognition.

\textbf{Slot Attention-based:} Slot Attention~\citep{locatello2020object} first initializes the representations of the object and iteratively updates them according to the similarity between the representations and the local features of the scene image. SAVi~\citep{kipf2022conditional} and SAVi++~\citep{elsayed2022savi++} use optical flow supervision to learn temporal information between object representations of adjacent frames. SLATE~\citep{singh2022illiterate} first encodes the image into latent, where it adopts Slot Attention with the difference that SLATE uses an auto-regressive transformer decoder. Based on the transformer decoder, STEVE~\citep{singh2022simple} makes significant improvements on various complex and naturalistic videos. DINOSAUR~\citep{Seitzer2022BridgingTG} adopts a pre-trained ViT model and performs slot attention at the latent scale. LSD~\citep{jiang2023object} adapts a pre-trained VQ-GAN encoder to transform the input image into latent and adds noise. Then the representations extracted by slot-attention are used as input to denoise the feature map. uORF~\citep{yu2021unsupervised} and ObSuRF~\citep{stelzner2021decomposing} decompose 3D scenes into objects via the Neural Radiance Fields (NeRFs)~\citep{mildenhall2021nerf} with the viewpoint annotations. While these Slot Attention-based methods excel at handling complex scenes, they fall short in entire scene generation and object identification across scenes.

It's important to note that, despite their strengths, both Non-Slot Attention and Slot Attention-based methods share a common limitation: they struggle with object identification tasks across scenes. GOCL \citep{chen2024compositional} is the only object-centric representation learning method to handle object identification tasks. However, GOCL can only handle simple 2D scenes. Therefore, it is still a huge challenge to identify objects across complex 3D scenes completely unsupervised.

\section{Method}
\label{sec:method}
\begin{figure}[tb]
  \centering
  \includegraphics[width=\columnwidth]{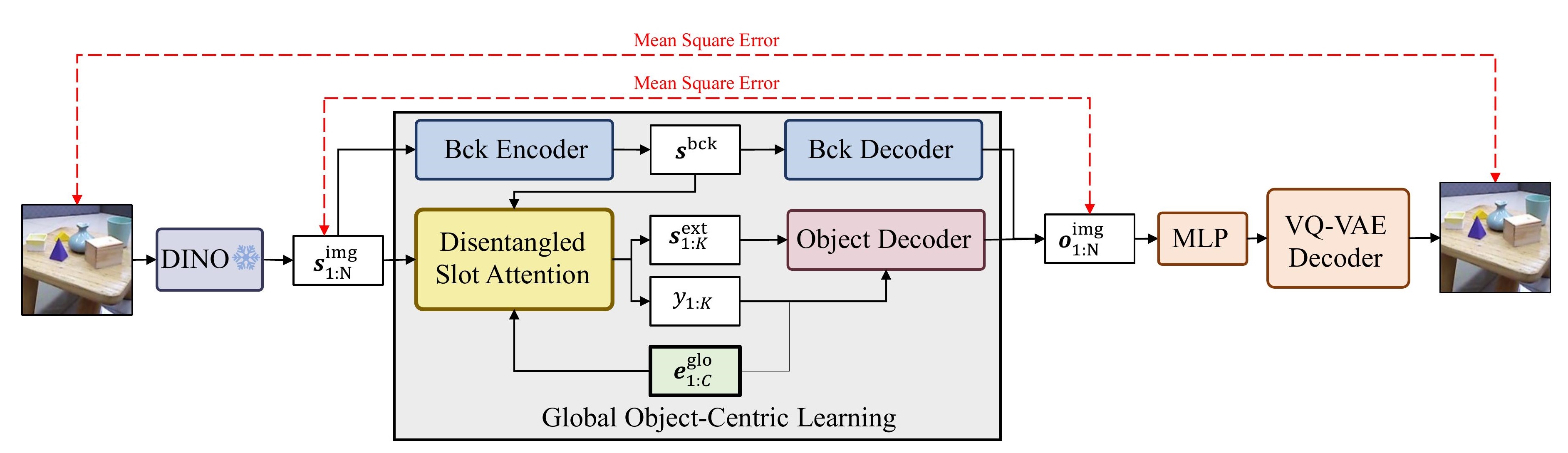}
  \caption{Overviews of GOLD. GOLD consists of the \textit{Image Encoder-Decoder} module and the \textit{Global Object-centric Learning} module. \textit{Image Encoder-Decoder} module is used to convert the input image into patch features and reconstruct the scene image via a VQ-VAE decoder with the reconstructed patch features. \textit{Global Object-Centric Learning} module is used to learn the global object-centric representations $\boldsymbol{e}_{1:C}^{\text{glo}}$ and extract the extrinsic representation $\boldsymbol{s}_{1:K}^{\text{ext}}$, identity representation $\boldsymbol{y}_{1:K}$ and background representation $\boldsymbol{s}^{\text{bck}}$. Background features are encoded and decoded individually using the \textit{Bck Encoder} and \textit{Bck Decoder}. \textit{Disentangled Slot Attention} module used to extract the scene-dependent attributes representation $\boldsymbol{s}_{1:K}^{\text{ext}}$ and scene-independent representation $\boldsymbol{y}_{1:K}$.  $\boldsymbol{y}_{1:K}$ is used to select the corresponding intrinsic representation from $\boldsymbol{e}_{1:C}^{\text{glo}}$, to reconstruct the patch features via \textit{Object Decoder}. \textit{ Image Encoder-Decoder} module is optimized by minimizing the Mean Square Error loss between the input and reconstructed images. \textit{Global Object-Centric Learning} module is trained by minimizing the Mean Square Error between extracted and reconstructed patch features.
  }
  \label{fig:overview}
\end{figure}

The main structure of the proposed method can be divided into two parts: an \textit{Image Encoder-Decoder} module and a \textit{Global Object-Centric Learning} module. As shown in Figure \ref{fig:overview}, firstly, the input image is converted into a patch features map through a pre-trained Transformer (DINO) \citep{kipf2022conditional}. Then, the patch features are fed into the \textit{Global Object-centric Learning} module to extract the extrinsic attributes representations, the identity representation and the reconstructed patch features map. Finally, the reconstructed patch features are used to reconstruct the input image via a VQ-VAE decoder \citep{van2017neural}. In the \textit{Global Object-Centric Learning} module, we introduce a \textit{Disentangled Slot Attention} module to disentangle the attributes of each object into intrinsic and extrinsic parts.

\subsection{Image Encoder-Decoder}
Similar to DINOSAUR~\citep{Seitzer2022BridgingTG}, SLATE~\citep{singh2022illiterate}, and LSD~\citep{jiang2023object}, we first transform the input image into a feature map and then extract the representations of object and background. This kind of design can help the model focus on features of a higher level when dealing with complex datasets. Given a static image $\boldsymbol{x} \in R^{H\times W \times C}$, we convert it into patch features $\boldsymbol {s}_{1:N}^{\text{img}} \in R^{N \times D_{\text{img}}}$ through a pre-trained Transformer DINO \citep{caron2021emerging} encoder. Different from DINOSAUR, we first extract the background representation from $\boldsymbol {s}_{1:N}^{\text{img}}$ and then extract the object representation based on the background representation and image patch features. To reconstruct the input image, we further input the reconstructed patch features $\boldsymbol{o}_{1:N}^{\text{img}}$ into a VQ-VAE Decoder $f_{\text{dec}}^{\text{VQ-VAE}}$ to get the reconstruction $\hat{\boldsymbol{x}}$. The complete process of \textit{Image Encoder-Decoder} module can be written as:
\begin{equation}
     \boldsymbol {s}_{1:N}^{\text{img}}=f_{\text{enc}}^{\text{img}}(\boldsymbol{x}), \quad 
     \boldsymbol {o}_{1:N}^{\text{img}} = f_{\text{gocl}}(\boldsymbol {s}_{1:N}^{\text{img}}), \quad
\hat{\boldsymbol{x}}=f_{\text{dec}}^{\text{VQ-VAE}}\Big(f_{\text{mlp}}^{\text{pat}}\big(\boldsymbol {o}_{1:N}^{\text{img}}\big)\Big),
\end{equation}
where $f_{\text{gocl}}$ denotes the \textit{Global Object-Centric Learning} module. $f_{\text{mlp}}^{\text{pat}}$ is a neural network and used to transform $\boldsymbol {o}_{1:N}^{\text{img}}$ into the same dimensions as the input of $f_{\text{dec}}^{\text{VQ-VAE}}$.

\begin{algorithm}[tb]
    \caption{Disentangled Slot Attention}
    \label{alg1}
 \begin{algorithmic}
     \State{\bfseries Input:} The patch features $\boldsymbol{s}^{\text{sce}}$ of input image, the background representation $\boldsymbol{s}^{\text{bck}}$, global object-centric representations $\boldsymbol{e}_{1:C}^{\text{glo}}$.
     \State{\bfseries Output:} The logits of the category distribution of object identity $\boldsymbol{\gamma}_{1:K, 1:C}$, the extrinsic attributes representations of the object $\boldsymbol{s}_{1:K}^{\text{ext}}$
     \State \textcolor{gray}{** Initialize the extrinsic attributes representation of the $k$th object**}
     \State$\boldsymbol{s}_{k}^{\text{ext}}\sim \mathcal{N}\big(\tilde{\boldsymbol{\mu}}^{\text{ext}},\text{diag}(\tilde{\boldsymbol{\sigma}}^{\text{ext}})^2\big), \forall 1  \leq   k  \leq K$ 
     \State \textcolor{gray}{** Initialize the logistics of the $k$th object**}
     \State $\boldsymbol{\gamma}_{k,1:C}\sim \mathcal{N}\big(\tilde{\boldsymbol{\mu}}^{\text{id}},\text{diag}(\tilde{\boldsymbol{\sigma}}^{\text{id}})^2\big), \forall 1  \leq   k  \leq K$ 
     \For{$t \leftarrow 1$ {\bfseries to} $T$}
     \State \textcolor{gray}{** Initialize the representation of the $k$th object identify**}
        \State $\boldsymbol{y}_{k} \leftarrow \text{Gumbel-Softmax}_{C}(\boldsymbol{\gamma}_{k,1:C}), 1\leq k \leq K $ 
        \State \textcolor{gray}{** Initialize the intrinsic attributes representation of the $k$th object**}
        \State $\boldsymbol{s}_{k}^{\text{int}} \leftarrow \sum_{c=1}^{C} y_{k,c}\cdot  f_{\text{glo}}(\boldsymbol{e}_{c}^{\text{glo}}), 1 \leq k \leq K$ 
        \State $\boldsymbol{s}_{k}^{\text{upd,int}} \leftarrow \boldsymbol{s}_{k}^{\text{int}}, \boldsymbol{s}_{k}^{\text{upd,ext}} \leftarrow \boldsymbol{s}_{k}^{\text{ext}}, 1 \leq k \leq K  $
        \State $\boldsymbol{s}_{k}^{\text{obj}} \leftarrow [\boldsymbol{s}_{k}^{\text{int}}, \boldsymbol{s}_{k}^{\text{ext}}], 1\leq k \leq K$
         \State $\boldsymbol{s}_{0:K}^{\text{full}} \leftarrow \big[\boldsymbol{s}_{1:K}^{\text{obj}},\boldsymbol{s}^{\text{bck}}\big]$
         \State$\boldsymbol{\tilde{a}}_{0:K}\leftarrow \text{softmax}_{K}\Big(\frac{1}{\sqrt{D_{\text{key}}}}\big(f_{\text{key}}(\boldsymbol{s}^{\text{sce}})\cdot f_{\text{qry}}(\boldsymbol{s}_{0:K}^{\text{full}})\big)\Big)$
         \State$\boldsymbol{w}_{0:K}\leftarrow \sum_{n=1}^{N}\text{Softmax}_{N}(\log \tilde{\boldsymbol{a}}_{n,0:K})\big)$
         \State$\boldsymbol{u}_{0:K}\leftarrow \boldsymbol{w}_{0:K} \cdot f_{\text{val}}(\boldsymbol{s}^{\text{sce}}\big)$
         \State $[\boldsymbol{u}_{1:K}^{\text{int}},\boldsymbol{u}_{1:K}^{\text{ext}}] \leftarrow \boldsymbol{u}_{1:K}$
         \State \textcolor{gray}{** Update the intrinsic attributes representation of the $k$th object**}
         \State$\boldsymbol{s}_{1:K}^{\text{upd,int}}\leftarrow f_{\text{GRU}}^{\text{int}}(\boldsymbol{s}_{1:K}^{\text{upd,int}},\boldsymbol{u}_{1:K}^{\text{int}})$
         \State \textcolor{gray}{** Update the extrinsic attributes representation of the $k$th object**}
         \State$\boldsymbol{s}_{1:K}^{\text{upd,ext}}\leftarrow f_{\text{GRU}}^{\text{ext}}(\boldsymbol{s}_{1:K}^{\text{upd,ext}},\boldsymbol{u}_{1:K}^{\text{ext}})$
         \State $\boldsymbol{s}_{1:K}^{\text{int}}\leftarrow \boldsymbol{s}_{1:K}^{\text{upd,int}}, \boldsymbol{s}_{1:K}^{\text{ext}}\leftarrow \boldsymbol{s}_{1:K}^{\text{upd,ext}}$
         \State \textcolor{gray}{** Update the logistics of the $k$th object**}
         \State$\boldsymbol{\gamma}_{k,1:C} \leftarrow \frac{1}{\sqrt{D_{\text{int}}}}\big(\boldsymbol{s}_{k}^{\text{int}}\cdot f_{\text{glo}}(\boldsymbol{e}_{1:C}^{\text{glo}})\big), 1\leq k \leq K$
     \EndFor
     \State\textbf{return} 
     $\boldsymbol{\gamma}_{1:K,1:C},\boldsymbol{s}_{1:K}^{\text{ext}}$
 \end{algorithmic}
 \end{algorithm}

\subsection{Global Object-Centric Learning}
\label{Global Object-Centric Learning}
To learn the scene-independent global object-centric representations, instead of extracting the scene-dependent object-centric representations of objects and backgrounds, we design to disentangle the representation of each object into extrinsic, intrinsic and identity representations. The extrinsic representation contains scene-dependent attributes, such as size, position and orientation, while the intrinsic representation captures scene-independent attributes, such as the globally invariant complete appearance and shape, and is selected from a set of learnable global object-centric representations according to the corresponding identity representation. Identity representation indicates the global identity of objects across scenes. Specifically, given the maximum number of objects in the scene $K$ and the number of objects types $C$, we extract three types of representations from each image feature map: background representations $\boldsymbol{s}^{\text{bck}}$, $K$ extrinsic  $\boldsymbol{s}_k^{\text{ext}}$ and identity $\boldsymbol{y}_k \in R^{C}$ representations of objects, where the range of the index $k$ is $1\leq k \leq K$ and will be omitted in following discussion for clarity. The extrinsic representations $\boldsymbol{s}^{\text{ext}}$ contains the spatial information(e.g., size, position, orientation) of objects while the identity representations $\boldsymbol{y}_k$ indicate the object prototypes which contain information that remains constant in the datasets (e.g., 3D shape and appearance). Before extracting $\boldsymbol{s}_k^{\text{ext}}$ and $\boldsymbol{y}_k$, we introduce $C$ learnable global object-centric representations $\boldsymbol{e}_{c}^{\text{glo}} (1 \leq c \leq C)$. The global object-centric representations store the information of total $C$ possible prototype objects appearing in all scenes and can be updated via the backpropagation algorithm.

We first encode the image feature map $\boldsymbol{s}_{1:N}^{\text{img}}$ into $\boldsymbol{s}^{\text{bck}}$ which then infers the mean $\boldsymbol{\mu}^{\text{bck}}$ and standard deviation $\boldsymbol{\sigma}^{\text{bck}}$ of the variational distribution of background latent variable $q(\boldsymbol{z}^{\text{bck}}|\boldsymbol{x})$. $\boldsymbol{z}^{\text{bck}}$ can be sampled from $q(\boldsymbol{z}^{\text{bck}}|\boldsymbol{x})$ by reparameterization trick \citep{kingma2014auto}, and fed into \textit{Bck Decoder} to get background appearance $\boldsymbol{a}_0 \in R^{N \times D_{\text{img}}}$ and its masks $\boldsymbol{m}_0 \in R^{N}$ through $f_{\text{enc}}^{\text{bck}}$ and $f_{\text{dec}}^{\text{bck}}$ respectively.
\begin{equation}
    \begin{aligned}
    \boldsymbol{s}^{\text{bck}} &=f_{\text{enc}}^{\text{bck}}(\boldsymbol{s}_{1:N}^{\text{img}}), \quad \quad 
    &\boldsymbol{\mu}^{\text{bck}}, \boldsymbol{\sigma}^{\text{bck}} &= f_{\text{lat}}^{\text{bck}}(\boldsymbol{s}^{\text{bck}}),  \\
    \boldsymbol{z}^{\text{bck}} &= \boldsymbol{\mu}^{\text{bck}} + \boldsymbol{\sigma}^{\text{bck}} \cdot \epsilon, \epsilon\sim\mathcal{N}(0,1) \quad \quad
    &[\boldsymbol{a}_0,\boldsymbol{m}_0] &=f_{\text{dec}}^{\text{bck}}(\boldsymbol{z}^{\text{bck}}), 
\end{aligned}
\end{equation}
where $f_{\text{enc}}^{\text{bck}}$, $f_{\text{lat}}^{\text{bck}}$ and $f_{\text{dec}}^{\text{bck}}$ are two-layer fully connected networks. 

The extracted patch features map $\boldsymbol{s}_{1:N}^{\text{img}}$, background representations $\boldsymbol{s}^{\text{bck}}$, and global object-centric representations $\boldsymbol{e}_{1:C}^{\text{glo}}$ are fed into a modified \textit{Disentangled Slot Attention} (DSA) module, which will be described in detail in the next section, to get extrinsic representations $\boldsymbol{s}_{1:K}^{\text{ext}}$ and the logits of the category distribution of object identity $\boldsymbol{\gamma}_{1:K, 1:C}$. $\boldsymbol{s}_{k}^{\text{ext}}$ is fed into a two-layer fully connected networks $f_{\text{lat}}^{\text{ext}}$ to infer the mean $\boldsymbol{\mu}_{k}^{\text{ext}}$ and standard deviation $\boldsymbol{\sigma}_{k}^{\text{ext}}$ of the variational distribution of extrinsic latent variable $q(\boldsymbol{z}_{k}^{\text{ext}}|\boldsymbol{x})$. $\boldsymbol{z}_{k}^{\text{ext}}$ can be obtained by $\boldsymbol{\mu}_{k}^{\text{ext}}$ and $\boldsymbol{\sigma}_{k}^{\text{ext}}$ via reparameterization trick. $\boldsymbol{\gamma}_{k,1:C}$ can be converted into the identity latent variable $\boldsymbol{y}_{k,1:C}$ by a Gumble-Softmax function \citep{Jang_Gu_Poole_2016}.
\vspace{-0.8em}
\begin{equation}
    \begin{aligned}
\vspace{-0.8em}
    (\boldsymbol{s}_{1:K}^{\text{ext}},\boldsymbol{\gamma}_{1:K, 1:C})&=\text{DSA}(\boldsymbol{s}_{1:N}^{\text{img}},\boldsymbol{s}^{\text{bck}},\boldsymbol{e}_{1:C}^{\text{glo}}),  
    &\boldsymbol{y}_{k,1:C} &= \text{Gumbel-Softmax}_{C}(\boldsymbol{\gamma}_{k,1:C}), \\
\boldsymbol{\mu}_{k}^{\text{ext}}, \boldsymbol{\sigma}_{k}^{\text{ext}} &= f_{\text{lat}}^{\text{ext}}(\boldsymbol{s}_{k}^{\text{ext}}), 
    &\boldsymbol{z}_{k}^{\text{ext}} &= \boldsymbol{\mu}_{k}^{\text{ext}} + \boldsymbol{\sigma}_{k}^{\text{ext}} \cdot \epsilon,
\end{aligned}
\end{equation}
where the range of index $k$ is $1\leq k \leq K$. $\epsilon$ is sampled from $\mathcal{N}(0,1)$.

For the $k$th object, we first compute its internal representations $\boldsymbol{s}_k^{\text{int}}$ by adding up $\boldsymbol{e}_{1:C}^{\text{glo}}$ with $\boldsymbol{y}_k$ as weights. Then we concatenate $\boldsymbol{s}_k^{\text{int}}$ and $\boldsymbol{z}_k^{\text{ext}}$ and input them into the \textit{Object Decoder} $f_{\text{dec}}^{\text{obj}}$ to get the corresponding appearance $\boldsymbol{a}_k$ and mask $\boldsymbol{m}_k$. Finally, we add up the appearances $\boldsymbol{a}_{0:K}$ with normalized masks $\hat{\boldsymbol{m}}_{0:K}$ as weights to get the reconstruction of image feature map $\boldsymbol{o}^{\text{img}}$. The complete process can be summarized as
\begin{equation}
    \begin{aligned}    \boldsymbol{s}_k^{\text{int}}&=\sum\nolimits_{c=1}^{C}\boldsymbol{e}_{c}^{\text{glo}} \cdot y_{k,c},
&\boldsymbol{a}_k,\boldsymbol{m}_k& =f_{\text{dec}}^{\text{obj}}\big(\left[\boldsymbol{s}_k^{\text{int}},\boldsymbol{z}_k^{\text{ext}}\right]\big),~1\leq k \leq K, \\
     \hat{\boldsymbol{m}}_{0:K}&=\text{softmax}_{K}(\boldsymbol{m}_{0:K}),\quad  &\boldsymbol{o}^{\text{img}}_{n}&=\sum\nolimits_{k=0}^{K}\boldsymbol{a}_k\cdot\hat{\boldsymbol{m}}_{k},~1\leq n \leq N.
\end{aligned}
\end{equation}

\textbf{Disentangled Slot Attention (DSA)} 
The original Slot Attention first initializes object representations and updates them through the cross-attention mechanism between the image feature map and initialized object representations. Therefore, the original slot attention can only extract scene-dependent object representations, making it impossible to achieve object identification across scenes. To address this problem, the proposed DSA divides object attributes into scene-dependent extrinsic and scene-independent intrinsic attributes. The extrinsic and intrinsic representations can be extracted from the whole scene patch features by defining a set of learnable global object-centric representations. 
 Similar to the SA, the extrinsic representation of each object in DSA is initialized by sampling from a Gaussian distribution with learnable parameters and then updated iteratively by the attention to the scene features. The intrinsic representation in DSA is one of the global object-centric representations selected according to the corresponding identity representation that is initialized and updated similarly to the extrinsic representation. By initializing and updating the extrinsic representation and identity representation separately, the extrinsic attributes of the object can be effectively disentangled from the intrinsic attributes, and the identity of the object can be accurately predicted.

Firstly, the extrinsic representation $\boldsymbol{s}_{1:K}^{\text{ext}}$ is initialized by sampling from a Gaussian distribution with learnable parameters $\boldsymbol{\tilde{\mu}}^{\text{ext}}$ and $\boldsymbol{\tilde{\sigma}}^{\text{ext}}$. The unnormalized probabilities $\boldsymbol{\gamma}_{1:K,1:C}$ of identity representation $\boldsymbol{y}_{1:K,1:C}$ is initialized by sampling from a Gaussian distribution with learnable parameters $\boldsymbol{\tilde{\mu}}^{\text{id}}$ and $\boldsymbol{\tilde{\sigma}}^{\text{id}}$. During each iteration process, for the $k$th object, we first sample its object identity representations $\boldsymbol{y}_k$ by Gumble-Softmax function \citep{Jang_Gu_Poole_2016} given current $\boldsymbol{\gamma}_k$. Then we compute the object representations $\boldsymbol{s}_k^{\text{obj}}$ given current $\boldsymbol{s}_k^{\text{ext}}$, $\boldsymbol{y}_k$, and $\boldsymbol{e}_{1:C}^{\text{glo}}$. We combine the $K$ object representations $\boldsymbol{s}_{1:K}^{\text{obj}}$ with the background representation as $K+1$ slots and compute $\boldsymbol{u}_{0:K}$ through the cross-attention mechanism similar to the original \textit{Slot Attention} algorithm. $\boldsymbol{u}_{1:K}$ is split into $\boldsymbol{u}_{1:K}^{\text{int}}$ and $\boldsymbol{u}_{1:K}^{\text{ext}}$, which will be input into GRU networks $\boldsymbol{f}_{\text{GRU}}^{\text{int}}$ and $\boldsymbol{f}_{\text{GRU}}^{\text{ext}}$ respectively to get updated object disentangled representations $\boldsymbol{s}_{1:K}^{\text{upd,int}}$ and $\boldsymbol{s}_{1:K}^{\text{upd,ext}}$. We directly treat $\boldsymbol{s}_{1:K}^{\text{upd,ext}}$ as $\boldsymbol{s}_{1:K}^{\text{ext}}$ for next iteration process. Finally, we compute the similarity between the $\boldsymbol{s}_{1:K}^{\text{int}}$ and $\boldsymbol{e}_{1:C}^{\text{glo}}$ to get updated $\boldsymbol{\gamma}_{1:K, 1:C}$. The object identity representation can be obtained by reparameterizing $\boldsymbol{\gamma}_{1:K, 1:C}$ via the Gumbel-Softmax function. The detailed process of the proposed DSA is shown in \ref{alg1}.

\subsection{Training}
The complete training process includes two parts: the feature reconstruction stage and the image reconstruction stage. In the first stage, only the \emph{Global Object-Centric Learning} module will be trained, which focuses on learning global object-centric representations, extrinsic representations, and identity representations.  It can get better object identification and scene decomposition performances by minimizing the feat reconstruction loss $\mathcal{L}_{\text{feat}}$ of the whole scene. In the second stage, only \emph{Image Encoder-Decoder} module will be trained, which focuses on reconstructing images of the individual object and the whole scene. It can obtain better image reconstruction and generation performance by minimizing the image reconstruction loss $\mathcal{L}_{\text{img}}$ of the whole scene. The total loss function $\mathcal{L}$ can be written as
\begin{equation}
   \mathcal{L}= \lambda_{\text{feat}}\mathcal{L}_{\text{feat}} + \lambda_{\text{img}}\mathcal{L}_{\text{img}},
\end{equation}
where $\mathcal{L}_{\text{feat}}$ and $\mathcal{L}_{\text{img}}$ stands for the loss function of the two stages. 

In order to obtain the best object identification and scene reconstruction and generation performance, the coefficients of the loss function corresponding to each stage in each stage of model training are set to 0. Specifically, in the first stage, the hyperparameter $\lambda_{\text{feat}}$ is set to 1, and $\lambda_{\text{img}}$ is set to 0, which can ensure that the model can learn better-disentangled representations. When the image feature reconstruction loss no longer decreases (In the actual implementation process, it can be replaced by setting the maximum number of training steps), the model enters the second stage of training. In the second stage, the hyperparameter $\lambda_{\text{feat}}$ is set to 0, and $\lambda_{\text{img}}$ is set to 1 to ensure that high-quality scene images can be obtained. When the image reconstruction loss no longer decreases (In the actual implementation process, it can also be replaced by setting the maximum number of training steps), the entire model training process ends. During the whole training process, we use linear warmup and cosine annealing to adjust the learning rate, and the temperature in the Gumble-Softmax function is also adjusted by cosine annealing.

\textbf{Feature reconstruction}
In the first stage, we mainly reconstruct the image feature map $\boldsymbol{s}_{1:N}^{\text{img}}$, and the corresponding loss function is
\begin{equation}
    \mathcal{L}_{\text{feat}}= \mathcal{L}_{\text{ELBO}} + \eta \frac{1}{N} \sum\nolimits_{n=1}^{N}{\big\Vert \boldsymbol{s}_{n}^{\text{img}} - \boldsymbol{a}_{0,n}\cdot\boldsymbol{\hat{m}}_{0,n} \big\Vert}^2_{2}. 
\end{equation}
The first term $\mathcal{L}_{\text{ELBO}}$ is the Evidence Lower Bound (ELBO). The second term is the regularization term, which measures the difference between the background feature map and the complete image feature map, encouraging the background representations to reconstruct more areas of the image feature map. All latent variables $\boldsymbol{\Omega} = \{\boldsymbol{z}^{\text{bck}}, \boldsymbol{z}^{\text{ext}},\boldsymbol{y}\}$ are inferred by amortized variational inference. The computation of $\mathcal{L}_{\text{ELBO}}$ can be further expanded as follows.
\begin{equation}
\begin{aligned}
    \mathcal{L}_{\text{ELBO}} = &-\mathbb{E}_{q(\boldsymbol{\Omega}|\boldsymbol{s}_{1:N}^{\text{img}},\boldsymbol{e}_{1:C}^{\text{glo}})}\big[\log p(\boldsymbol{s}_{1:N}^{\text{img}}|\boldsymbol{\Omega};\boldsymbol{e}_{1:C}^{\text{glo}})\big]\!\!+\!\!D_{\text{kl}}\Big(q\big(\boldsymbol{z}^{\text{bck}}|\boldsymbol{s}_{1:N}^{\text{img}})||p(\boldsymbol{z}^{\text{bck}}\big)\Big)\\
  &+ \sum\nolimits_{k=1}^{K}D_{\text{kl}}\Big(q\big(\boldsymbol{z}_{k}^{\text{ext}}|\boldsymbol{s}_{1:N}^{\text{img}}, \boldsymbol{z}^{\text{bck}};\boldsymbol{e}_{1:C}^{\text{glo}})||p(\boldsymbol{z}_{k}^{\text{ext}}\big)\Big) \\
&+\sum\nolimits_{k=1}^{K}D_{\text{kl}}\Big(q\big(y_{k}|\boldsymbol{z}_{k}^{\text{ext}},\boldsymbol{s}_{1:N}^{\text{img}};\boldsymbol{e}_{1:C}^{\text{glo}})||p(y_{k}\big)\Big) 
\end{aligned}
\end{equation}

\textbf{Image reconstruction} In the second stage, we also reconstruct the input image $\boldsymbol{x}$ to enable the model to generate pixel-level images. It can be written as
\begin{equation}
   \mathcal{L}_{\text{img}}= {\big\Vert \boldsymbol{x}-\hat{\boldsymbol{x}}\big\Vert}^2_{2}. 
\end{equation}


\section{Experiments}
\label{sec:experiments}
In this section, GOLD is evaluated on four datasets, including three synthetic scene datasets and one real-world scene dataset. The experimental results verify GOLD's object identification capabilities, global object-centric representation learning capabilities, and the ability to disentangle object attributes. Furthermore, we demonstrate the superiority of GOLD in individual object generation and scene decomposition.

\textbf{Datasets}
We conducted experiments on four datasets: CLEVR~\citep{johnson2017clevr}, SHOP~\citep{nazarczuk2020shop}, GSO~\citep{downs2022google} and OCTScene-A (OCTA)~\citep{huang2023octscenes}. The first three datasets are synthetic, whereas the fourth is a real-world dataset. In contrast to the standard CLEVR dataset, we expanded the variety of objects from 3 to 10, introducing more complex shapes. The SHOP dataset consists of 10 distinct object types against a single background, featuring textures of objects that display greater complexity compared to CLEVR. The GSO dataset, generated by Kubric~\citep{greff2022kubric}, encompasses 10 object types and 10 background variations, both exhibiting a higher level of texture complexity compared to SHOP and CLEVR. OCTA is a real-world tabletop scene dataset featuring 11 object types and a single background. Unlike CLEVR, SHOP, and GSO, where the number of objects in each scene is from 3 to 6, OCTA  is from 1 to 6.

\textbf{Baseline}
Three object-centric learning methods are selected as baselines: STEVE, DINOSAUR, and LSD. STEVE is a representative approach employing a transformer-based decoder for complex and naturalistic videos. Aligning with other comparison methods, STEVE is trained and tested on single images instead of videos. DINOSAUR, akin to GOLD, uses DINO for feature extraction and reconstructs features. LSD is chosen for its outstanding slot-to-image decoding quality achieved by integrating diffusion models. In addition, GOCL is regarded as a baseline because it can identify the same object across scenes like GOLD.

\textbf{Evaluation Metrics} 
Several metrics are used to evaluate the performance of scene decomposition. \emph{Adjusted Rand Index} (ARI) \citep{hubert1985comparing}, and \emph{mean Intersection over Union} (mIOU) are used to assess the quality of segmentation. ARI-A is computed considering all pixels in the scene, while ARI-O is calculated considering only the pixels of objects. The better the performance, the higher the value of ARI and mIOU. Object identification performance is measured by \emph{Accuracy} (ACC),  indicating the ability to predict object global identity across scenes. All the reported metrics are based on three evaluations.

\textbf{Implementation details} 
We train the model on a single GeForce RTX 3090 GPU with a mini-batch size of 8. The total number of steps is $300K$. The initial learning rate is $4\times 10^{-5}$ and $3\times 10^{-4}$ for \textit{Global Object-Centric Learning} and \textit{Image Encoder-Decoder} module respectively. They all decayed exponentially with a factor of 0.5 every 100,000 steps after multiplying a parameter that increases linearly from 0 to 1 in the first 10000 steps. The number of slots is 8 (including seven objects and one background) for all datasets. The number of patches of scene features is 768.  The size of object representations is 64 for the CLEVR and SHOP datasets and 128 for the GSO and OCTA datasets. The size of background representations is 4 for the CLEVR and SHOP and 8 for the GSO and OCTA datasets. $\eta$ is 0.001 for the CLEVR and GSO datasets, 0.01 for the SHOP dataset, and 0.1 for the OCTA dataset. $C$ is 10 for the CLEVR, SHOP and GS datasets and 11 for the OCTA dataset.

\subsection{Object Identification}
Except for GOCL, other object-centric learning methods have no object identification capability. We measure the object identification performance of GOLD by the accuracy (ACC) of predicting the global identity of objects across the scenes. The ACC of GOLD and GOCL can be calculated by matching the predicted identity representation with the identity in the label. The accuracy of other comparative methods, such as DINOSAUR, LSD, and STEVE, is calculated by first clustering the extracted object-centric representations and then matching the output of clustering with the object identity in the label. The object identification results of GOLD, GOCL, DINOSAUR, LSD and STEVE on the CLEVR, SHOP, and GSO datasets are shown in Table~\ref{tab:acc}. 

As can be seen from Table~\ref{tab:acc}, the proposed method achieves significantly better object identification performance than the comparison methods on all three datasets, which means that the identity representation extracted by GOLD can effectively predict the global identity of each object in all complex 3D scenes. Although GOCL has the ability to identify objects across scenes, the ACC of GOCL on the three datasets is much lower than GOLD and even the object-centric learning methods without object identification ability, such as DINOSAUR, LSD, and STEVE. This is because GOCL is only suitable for simple 2D scenes and cannot handle complex 3D scenes. It can be proved by the results in Figure~\ref{fig:seg}. It shows that GOCL cannot even decompose complex 3D scenes well, which may result in the inability to identify objects.

It is worth noting that although the textures of objects and backgrounds in the GSO dataset are the most complex of the three datasets, GOLD has the best object identification performance on the GSO dataset. The possible reason is that most of the shapes of objects in the GSO dataset are relatively regular and convexly symmetrical. For example, medicine bottles, pencil cases, etc. are cylinders, and tapes are rings, etc. The difference in the shape of the same object observed in different scenes is relatively small, which makes it easier to correctly identify the object across scenes. On the contrary, the shapes of most objects in the CLEVR and SHOP datasets are seriously irregular, such as horses, ducks, and teapots in CLEVR, and kettles and pans in the SHOP dataset. The difference in the shape of the same object observed in different scenes is large, making it difficult to identify objects across scenes. GOCL also has similar results to GOLD. However, DINOSAUR, LSD, and STEVE show different results. This is because DINOSAUR, LSD, and STEVE can only predict object identities by clustering scene-dependent object representations. Compared with CLEVR and SHOP, the scenes in the GSO dataset are more complex, and the extracted object-centric representations are more related to the scene, resulting in lower ACC.

\begin{table*}[t]
\caption{\textbf{The comparison results of object identification performance (ACC)} of GOLD, GOCL, DINOSAUR, LSD, and STEVE on the CLEVR, SHOP, and GSO datasets. }
\label{tab:acc}
\centering
\footnotesize
\begin{tabular}{lcccccc}
\toprule
\bfseries DATASET &\bfseries GOLD &\bfseries GOCL &\bfseries DINOSAUR &\bfseries LSD &\bfseries STEVE  \\
\midrule 
  CLEVR &\bfseries 0.766$\pm$8e-4 &0.370$\pm$3e-3 &0.691$\pm$2e-3 &0.705$\pm$8e-3 &0.565$\pm$2e-2\\ 
  SHOP  &\bfseries 0.804$\pm$2e-3 &0.369$\pm$6e-3 &0.684$\pm$3e-4 &0.709$\pm$1e-2 &0.686$\pm$6e-2\\
  GSO   &\bfseries 0.852$\pm$1e-3 &0.444$\pm$1e-2 &0.663$\pm$3e-3 &0.655$\pm$3e-3 &0.333$\pm$3e-3\\  
\bottomrule
\end{tabular}
\vskip -0.15in
\end{table*}

\subsection{Global Object-Centric Representations Learning}
In order to comprehensively showcase the globally learned object-centric representation by GOLD, we visualize the acquired global object-centric representation from two perspectives: prototype images and the scene generation with specific objects.

\begin{wrapfigure}{r}{0.7\columnwidth}
\vspace{-0.25in}
  \centering
  \includegraphics[width=0.7\columnwidth]{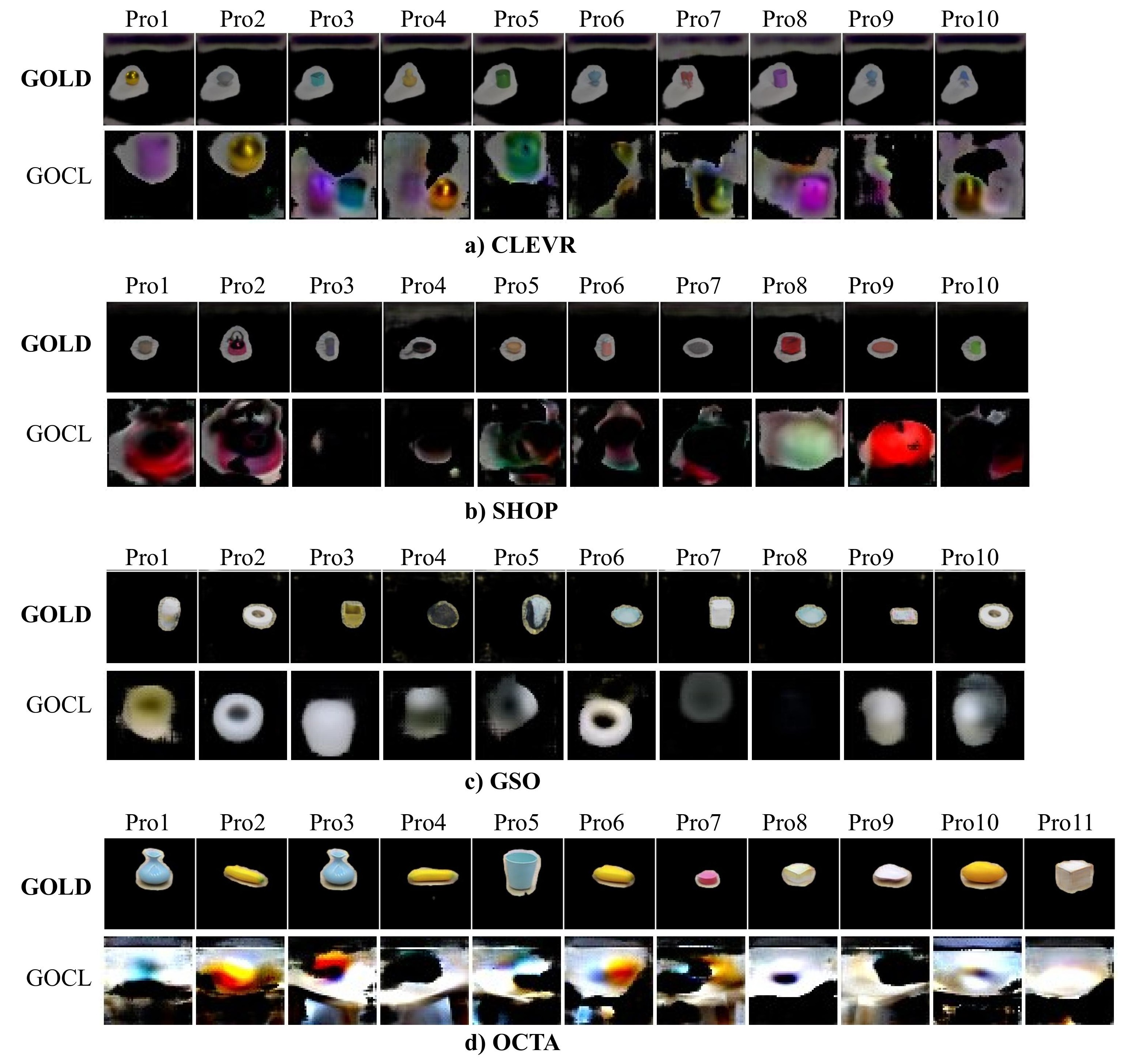}
  \vskip -0.08in
  \caption{\textbf{The visualization of Prototype Images} generated by GOLD and GOCL. From `Pro1' to `Pro10'(or `Pro11'), denote one prototype image generated, respectively.
  }
  \label{fig:protos}
  \vskip -0.25in
\end{wrapfigure}

\textbf{Prototype Images} Each prototype image can be generated through a global object-centric representation together with an extrinsic attribute representation. A visualization of the prototype image generated by GOLD and GOCL on four datasets is shown in Figure~\ref{fig:protos}. From Figure~\ref{fig:protos}, we can see that GOLD can generate images of almost all object prototypes on all four datasets, which means that GOLD can effectively learn satisfactory global object-centric representations from the four datasets. In contrast, GOCL can only generate a few object prototype images or even fail to generate complete images. This implies that GOCL cannot learn global object-centric representations from complex 3D scenes.

\begin{figure}[th]
 \vspace{-0.18in}
  \centering
  \includegraphics[width=0.95\columnwidth]{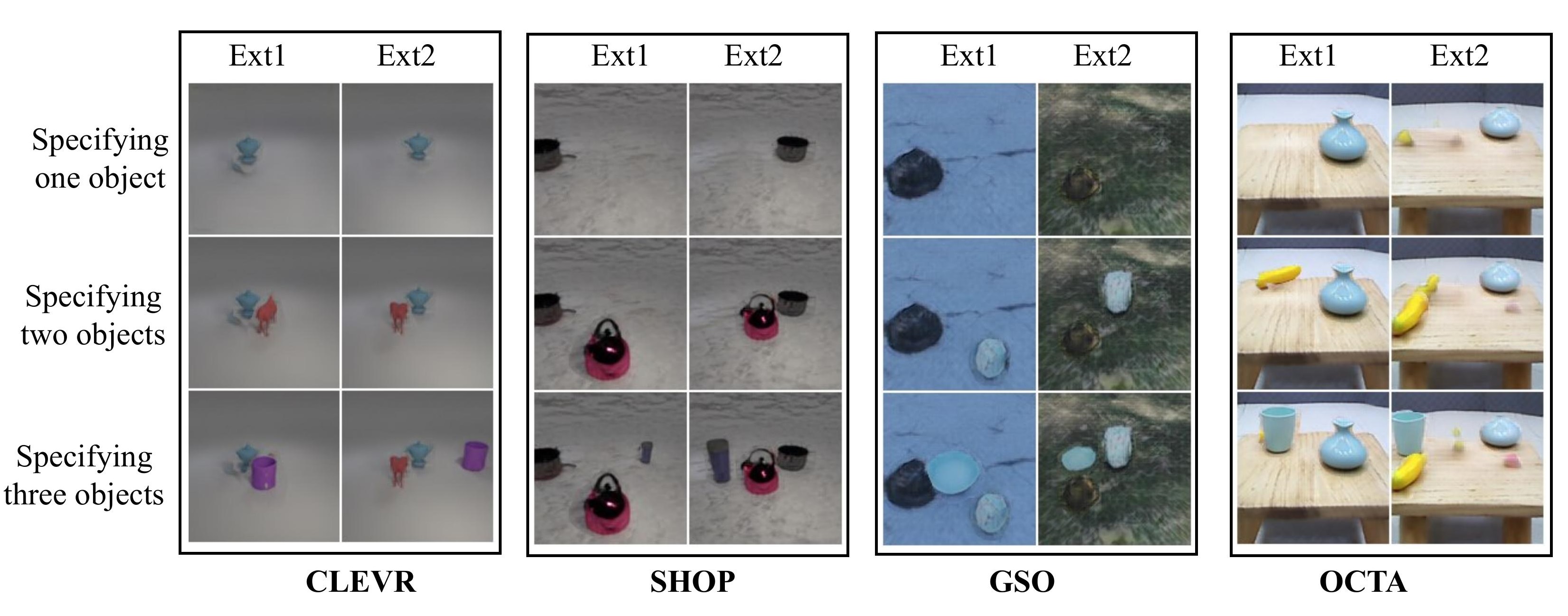}
  \caption{\textbf{The visualization of Scene Generation with Specific Objects} on the GSO dataset. `Ext1' and `Ext2' indicate two different representations of the extrinsic attributes. `Sample1'$\sim$`Sample4' denote four samples each of which contains six scene images. In each sample, each row of scene images contains the same objects, and each column contains increasing numbers of objects. The global object-centric representations of the two columns of scene images are the same, but the extrinsic attribute representations are different.
  }
  \label{fig:spec}
   \vspace{-0.236in}
\end{figure}

\textbf{Scene Generation with Specific Objects} To verify that the global object-centric representation learned by GOLD contains only scene-independent attributes (i.e., appearance and shape) but not scene-dependent attributes (i.e., size, position, orientation, etc.), we generate scenes containing specific objects by specifying object identity and object extrinsic attribute representation separately. Figure~\ref{fig:spec} shows the visualization of a scene generated by GOLD, specifying multiple objects. Two samples of scenes with different extrinsic attributes of four datasets containing 1$\sim$3 objects generated by GOLD are shown in Figure~\ref{fig:spec}. The type and number of objects in each sample (column) are consistent, and the size, position, and orientation of identical objects across different rows within each column of images are also consistent. The number of objects and intrinsic attributes of the two columns in the same row are the same, but the extrinsic attributes differ. For example, the position and orientation of the red horse in the first column (`Ext1') and the second column (`Ext2') of the CLEVR dataset are different. In addition, the size, position and orientation of the teapot in the SHOP dataset, the school bag in the GSO dataset, and the banana in the OCTA dataset are different in the first and second columns of the scene. Therefore, it can be shown that GOLD can accurately control the type and number of objects in the scene through learned global object-centric representations. Furthermore, the size, position, and orientation of each object in the scene can be controlled via the extrinsic attribute representation of objects.

\subsection{Attributes Disentanglement }
To verify the ability of GOLD to disentangle object attributes, we first select two samples from each dataset (`sample1' and `sample2'). Subsequently, we choose two objects in each scene to exchange their extrinsic attribute representations. Finally, we generate the scene image (`AEE') containing the objects with exchanged extrinsic attribute representations and compare it with the reconstructed scene image (`BEE' ) before exchanging extrinsic attribute representations to evaluate the difference.

\begin{figure}[th]
\vspace{-0.2in}
  \centering
  \includegraphics[width=\columnwidth]{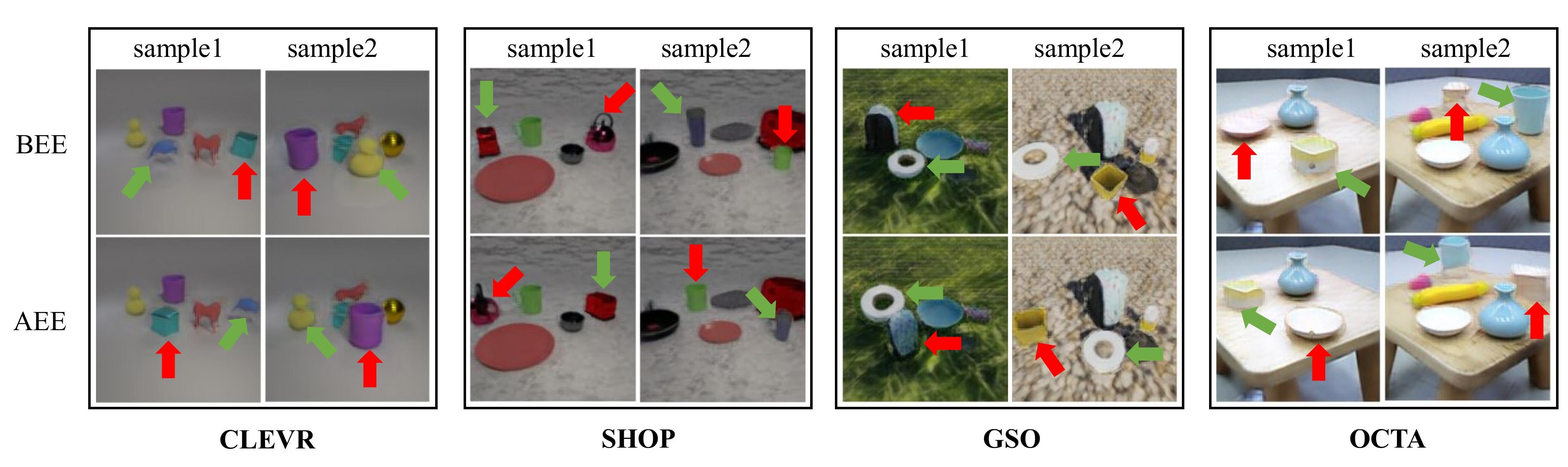}
  \vspace{-0.2in}
  \caption{\textbf{The visualization of Attributes Disentanglement } of GOLD on four datasets. `BEE' indicates the reconstructed scene before exchanging extrinsic attributes. `AEE' denotes the generated scene after exchanging extrinsic attributes. Each dataset showed two samples (`sample1' and `sample2'). The red and green arrows in each sample point to the two objects whose extrinsic attributes are exchanged. In the `BEE' and `AEE' rows, the arrows of the same color point to the objects with the same intrinsic attributes, i.e., the two objects before and after exchanging the extrinsic attributes.
  }
  \label{fig:disen}
  \vspace{-0.2in}
\end{figure}

The visualization results of GOLD's disentanglement of object attributes on four datasets are shown in Figure~\ref{fig:disen}. In each sample, the size, position, and orientation of the two objects in the first row (`BEE') scene changed significantly in the generated scene image (`AEE') after exchanging extrinsic attribute representations. For example, in the first sample of the SHOP dataset, after the kettle and the bread oven in the `BEE' scene exchanged extrinsic attribute representations, the position of the kettle changed from the far right to the far left, and the orientation of the kettle also changed significantly. In the first sample of the GSO dataset, after the school bag and the white tape in the `BEE' scene exchanged extrinsic attributes, it was found that the orientation of the school bag in `AEE' changed significantly, the size also became smaller, and the position was further down. In summary, the visualization results in Figure \ref{fig:disen} demonstrate GOLD's excellent object attribute disentanglement capabilities.

\subsection{The Individual Object Generation}
\begin{wrapfigure}{r}{0.7\columnwidth}
\vspace{-0.25in}
  \centering
  \includegraphics[width=0.7\columnwidth]{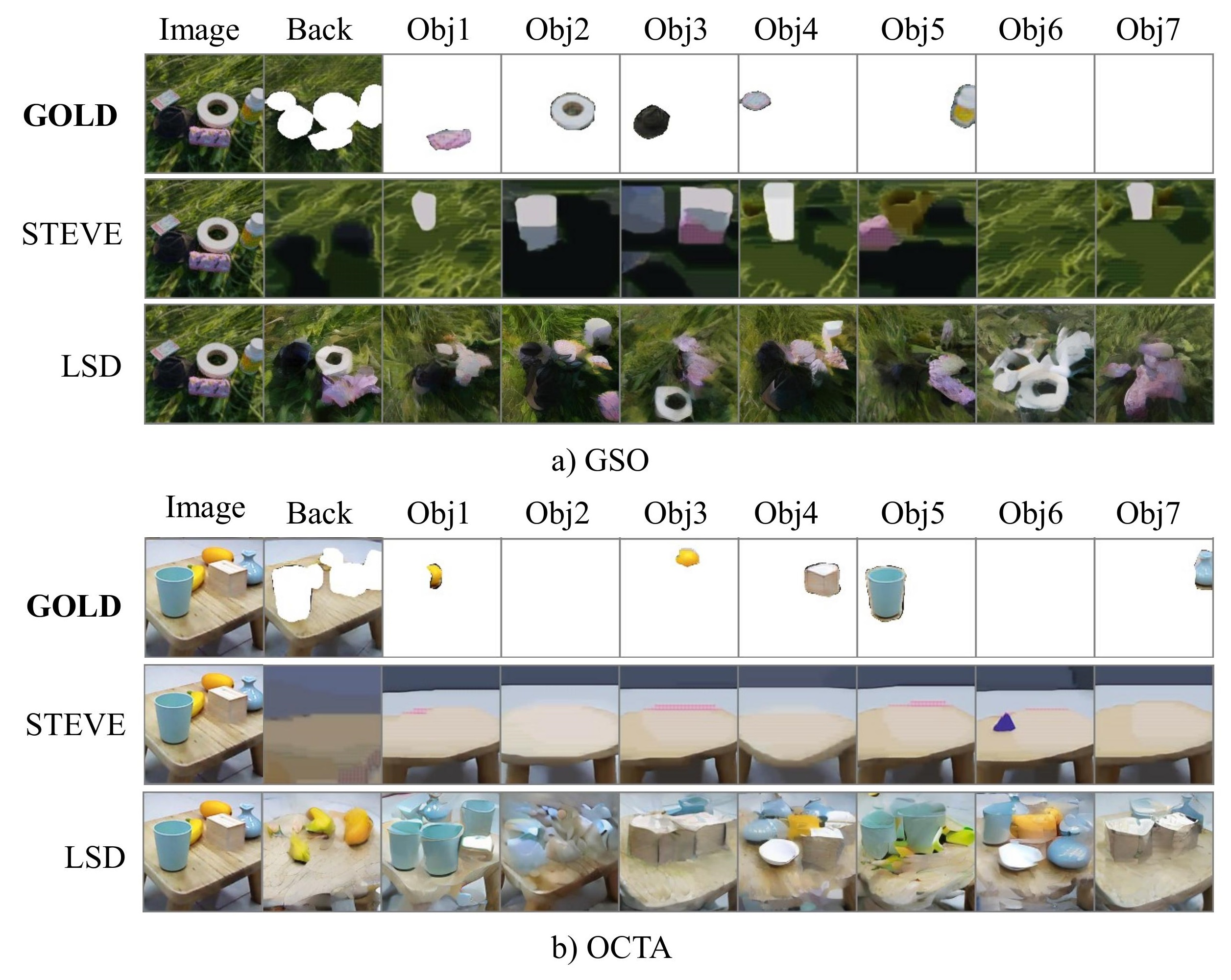}
  \vspace{-0.2in}
  \caption{\textbf{The visualization of Individual Object Generation} of GOLD, STEVE and LSD on the GSO and OCTA datasets. `Image' indicates the input image. `Back' denotes the background generated via the background representation (Only GOLD). `Obj1' $\sim$`Obj8' presents the individual object image generated via the corresponding extrinsic representation and the selected global object-centric representation.
  }
  \label{fig:indobj}
  \vskip -0.22in
\end{wrapfigure}

In this section, we evaluated the models' ability to generate images containing only a single object, conditioned on the representation of that object. The comparison of generation results for individual objects on the GSO and OCTA datasets is illustrated in Figure \ref{fig:indobj}. From the visualization results, it is evident that only GOLD successfully generates images containing a single object, whereas STEVE and LSD fail to achieve this objective. When conditioned on a single object representation, STEVE and LSD produce chaotic images that retain the background of the original image and may contain fewer or more objects. This limitation arises from the fusion of information between slots during decoding for the reconstruction of the entire image, leading to a reduction in the independence of object representation. Therefore, the images generated by these two methods contain information about objects other than the specified ones. In contrast, GOLD establishes a one-to-one correspondence between a single object representation and the generation of images containing only that object through a mixture-based decoder. This approach balances the impact of reconstructing the entire scene and reconstructing a single object by prioritizing the characteristics of the individual object.

\subsection{Scene Decomposition}

\begin{figure}[tb]
  \centering
  \includegraphics[width=\columnwidth]{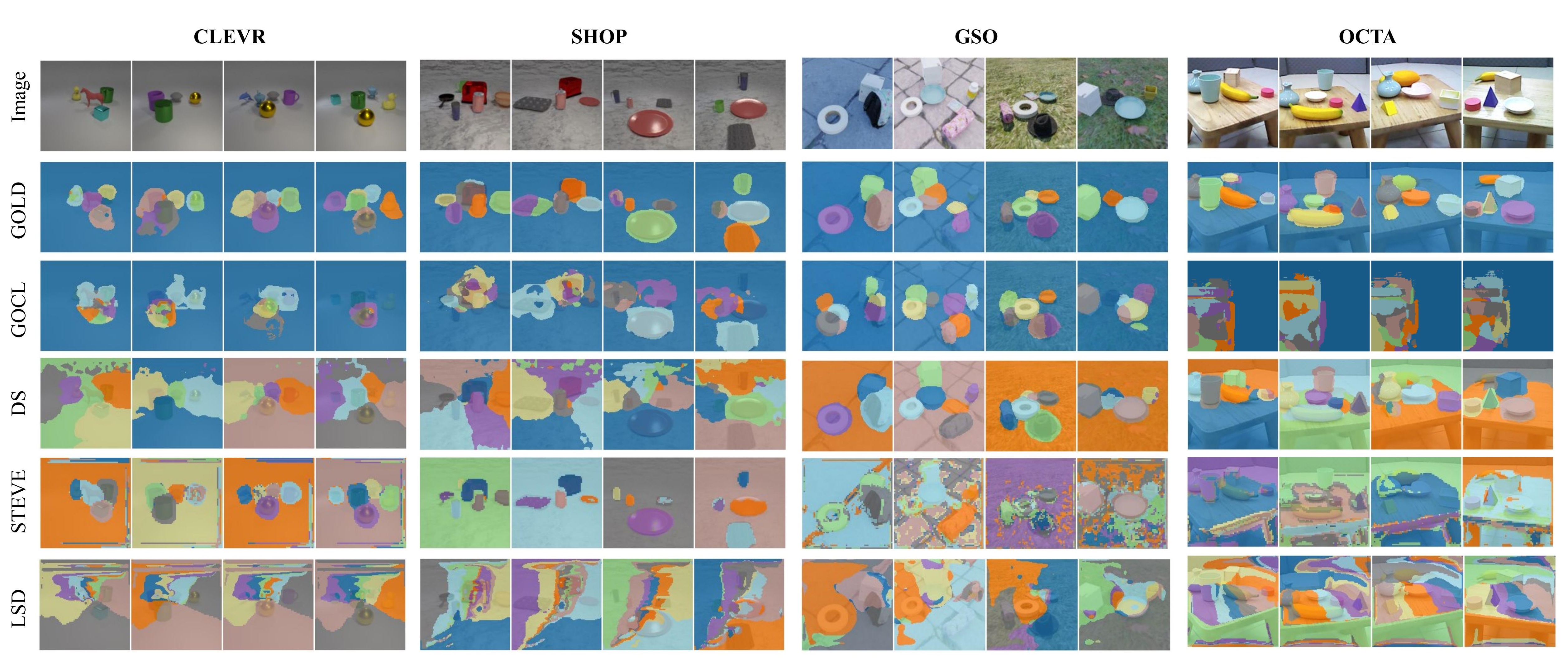}
  \vspace{-0.2in}
  \caption{\textbf{The visualization of unsupervised scene segmentation} of GOLD, GOCL, DINOSAUR, STEVE, and LSD on four datasets. `DS' indicates DINOSAUR. `Image' denotes the input image.  }
  \label{fig:seg}
  \vskip -0.1in
\end{figure}

We show the visualization of the scene decomposition performance of GOLD and baselines on four datasets in Figure~\ref{fig:seg}. GOLD demonstrates accurate segmentation of all objects and backgrounds across all datasets, showcasing superior scene understanding compared to the comparison methods. Although GOCL can segment the background relatively completely like GOLD, it is difficult for GOCL to segment the foreground objects accurately, and it often divides the foreground objects into the background. This shows that GOCL is almost unable to handle complex 3D scenes. Although DINOSAUR can partition different objects, its segmentation on the CLEVR and SHOP datasets is rough. Despite the segmented object mask not including other objects, DINOSAUR frequently mislabels substantial background areas as parts of the object, resulting in segmentation masks often being significantly larger than the actual mask of the object. While STEVE performs well on the CLEVR and SHOP datasets, its performance is significantly inferior to GOLD's on the more complex synthetic dataset GSO and the real-world dataset OCTA. Specifically, in GSO, STEVE occasionally segments two objects into one, as seen with the pencil case and medicine bottle both being labeled as orange masks in the second column. Moreover, OCTA presents great challenges for STEVE, making accurate foreground object segmentation almost unattainable. Similar to DINOSAUR, LSD also demonstrates rough segmentation, frequently segmenting background elements as foreground objects. Additionally, LSD's segmentation performance notably falls short of GOLD's.

The quantitative segmentation results of GOLD and the comparison methods on four scenes are shown in Table~\ref{tab:metrics}. It can be seen from the table that in terms of foreground object segmentation (ARI-O), except for DINOSAUR, GOLD is significantly better than other comparison methods on all datasets. This shows that GOLD has satisfactory segmentation capabilities. It is worth noting that in terms of the segmentation metric (ARI-A) considering both the object and background, GOLD is significantly better than DINOSAUR on most data sets, thanks to GOLD's special modeling of the background so that objects and background in the scene can be distinguished. In addition, we noticed that GOCL has woefully inadequate segmentation performance on all datasets, which means that GOCL can hardly handle complex 3D scenes.

\begin{table*}[t]
    \centering
    \footnotesize
    \caption{The quantitative segmentation comparison results of GOLD, GOCL, DINOSAUR, STEVE and LSD in terms of all metrics on four datasets. }\label{tab:metrics}
    \begin{tabular}{lcccc}
      \toprule 
      \bfseries Data set & \bfseries Model &\bfseries ARI-A$\uparrow$ &\bfseries ARI-O$\uparrow$ &\bfseries mIOU$\uparrow$ \\
      \midrule 
      \multirow{5}{*}{CLEVR} 
    &GOLD         &\underline{0.362$\pm$5e-3} &\underline{0.904$\pm$4e-3} &\bfseries 0.425$\pm$8e-4 \\
    &GOCL         &\bfseries0.403$\pm$3e-3 &0.076$\pm$2e-3 &0.067$\pm$5e-4 \\
    &DINOSAUR     &0.038$\pm$2e-3 &\bfseries 0.956$\pm$4e-3 &0.186$\pm$1e-3 \\
    &STEVE        &0.303$\pm$1e-3 &0.878$\pm$1e-2 &\underline{0.373$\pm$1e-3} \\
    &LSD          &0.045$\pm$4e-4 &0.516$\pm$6e-4 &0.157$\pm$9e-4\\
      \midrule
      \multirow{5}{*}{SHOP} 
    &GOLD         &\underline{0.622$\pm$2e-3} &\underline{0.881$\pm$3e-3} &\underline{0.542$\pm$8e-3} \\
    &GOCL         &0.381$\pm$5e-3 &0.111$\pm$6e-3 &0.006$\pm$7e-5 \\
    &DINOSAUR     &0.032$\pm$2e-3 &\bfseries 0.946$\pm$3e-5 &0.212$\pm$1e-3 \\
    &STEVE        &\bfseries 0.836$\pm$1e-3 &0.783$\pm$1e-4 &\bfseries 0.703$\pm$4e-3 \\
    &LSD          &0.029$\pm$1e-4 &0.436$\pm$1e-3 &0.136$\pm$5e-4  \\
    \midrule
      \multirow{5}{*}{GSO} 
    &GOLD         &\bfseries0.693$\pm$2e-4 &\underline{0.912$\pm$3e-3} &\bfseries0.658$\pm$4e-4 \\
    &GOCL         &0.471$\pm$3e-2 &0.659$\pm$2e-3 &0.107$\pm$2e-4 \\
    &DINOSAUR     &\underline{0.689$\pm$1e-3} &\bfseries 0.928$\pm$3e-3 &\underline{ 0.632$\pm$1e-3} \\
    &STEVE        &0.235$\pm$2e-4 &0.361$\pm$1e-4 &0.244$\pm$1e-3 \\
    &LSD          &0.126$\pm$2e-3 &0.719$\pm$6e-3 &0.263$\pm$3e-3 \\ 
    \midrule
      \multirow{5}{*}{OCTA} 
    &GOLD         &\bfseries 0.722$\pm$4e-3 &\underline{0.706$\pm$2e-3} &\bfseries 0.528$\pm$5e-3 \\
    &GOCL         &0.048$\pm$3e-3 &0.031$\pm$2e-3 &0.042$\pm$1e-4 \\
    &DINOSAUR     &\underline{0.290$\pm$2e-3} &\bfseries0.759$\pm$1e-2 &\underline{0.462$\pm$5e-3} \\
    &STEVE        &0.001$\pm$1e-4 &0.138$\pm$2e-4 &0.094$\pm$2e-4 \\
    &LSD          &0.056$\pm$7e-5 &0.380$\pm$2e-4 &0.177$\pm$2e-4 \\
    \bottomrule 
    \end{tabular}
    \vskip -0.15in
\end{table*}

\subsection{Ablation Study}
In order to verify the importance of the proposed DSA module and learnable global object-centric representations to the object identification ability of GOLD, we designed two ablation models: `w/o DSA' and `w/o GLO'.   The original slot attention module is used instead of the proposed DSA in the `w/o DSA' model to verify the impact of the proposed DSA on the object identification performance of GOLD. `w/o GLO' denotes the model without defining the global object-centric representations and with simple modifications to the DSA and Object Decoder: 1) The extrinsic attribute representation and identity representation (instead of the intrinsic representation selected from the global object-centric representations) are directly updated in the DSA module; 2) The input of the Object Decoder becomes the extrinsic attribute representation and the identity representation (rather than the intrinsic attribute representation selected from the global object-centric representations).

The comparison results of the three models are shown in Table~\ref{tab:abla}. As can be seen from Table~\ref{tab:abla}, the object identification performance of the `w/o DSA' and `w/o GLO' models on the three datasets is significantly lower than that of GOLD, indicating that the proposed DSA and learnable global object-centric representations have a significant impact on the object identification ability of GOLD. It is noted that the object identification accuracy of `w/o GLO' on the three datasets is higher than that of `w/o DSA.'  This implies that the proposed DSA is more critical for the object identification ability of GOLD.

\begin{table*}[t]
\caption{\textbf{Ablation study results of GOLD, `w/o DSA' and `'w/o GLO' model in ACC metric on three datasets.} `w/o DSA' denotes the model without DSA. 'w/o GLO' indicates the model without global object-centric representations.  }
\label{tab:abla}
\centering
\footnotesize
\begin{tabular}{lccc}
\toprule
\bfseries Model &\bfseries CLEVR &\bfseries SHOP &\bfseries GSO  \\
\midrule 
  w/o DSA  &0.369$\pm$4e-4 &0.395$\pm$3e-3 &0.381$\pm$4e-3 \\ 
  w/o GLO  &0.383$\pm$3e-3 &0.484$\pm$4e-4 &0.475$\pm$2e-3 \\
  GOLD     &\bfseries 0.766$\pm$8e-4 &\bfseries 0.804$\pm$2e-3 &\bfseries 0.852$\pm$1e-3 \\  
\bottomrule
\end{tabular}
\vskip -0.15in
\end{table*}

\subsection{Limitation and Future Works}
The limitation of GOLD lies in the need to enhance the quality of reconstructed and generated images. Although our method demonstrates richer image generation capabilities, there is still room for improvement in image generation quality due to the relatively simple decoder employed, which fails to capture the complex texture of the images. In future work, we aim to utilize diffusion models with more robust generative capabilities as decoders to enhance the quality of scene reconstruction and generation.
\section{Conclusion}
\label{sec:conclusion}
In this paper, we propose an object-centric learning method called  \textbf{G}lobal \textbf{O}bject-centric \textbf{L}earning via \textbf{D}isentangled slot attention (GOLD), which acquires the ability to identify objects across various scenes and generate diverse scenes containing specific objects through a set of learnable global object-centric representations. To better identify objects in a scene, we first extract representations of the background and then use the extracted background with the input scene to extract representations of foreground objects. The \textit{Disentangled Slot Attention} module is introduced to convert scene features into scene-dependent and scene-independent attribute representations for each object in the scene. Experimental results on four datasets demonstrate that GOLD significantly outperforms other comparative methods in object identification and scene generation with specific objects in global object-centric representation learning.

\newpage

\bibliography{sn-article}

\end{document}